\documentclass[conference,a4paper]{IEEEtran}

\usepackage{amsmath,amssymb,amsthm}
\usepackage[ruled,vlined]{algorithm2e}
\usepackage{graphicx}
\usepackage{xcolor}
\usepackage{booktabs}
\usepackage{multirow}
\usepackage{listings}
\usepackage{url}
\usepackage[hidelinks]{hyperref}
\usepackage{enumitem}
\usepackage{tikz}
\usetikzlibrary{arrows.meta,positioning,shapes.geometric}

\IEEEoverridecommandlockouts
\IEEEpubid{\makebox[\columnwidth]{979-8-3195-3997-7/26/\$31.00~\copyright2026 IEEE\hfill}\hspace{\columnsep}\makebox[\columnwidth]{}}

\title{MakSafe: A Spatiotemporal Semantic Framework for Formal Specification and Compliance Checking of Road Vehicle Safety Rules}

\author{%
\IEEEauthorblockN{Maksym Labzhaniia, Julien Deantoni, Marie-Agn\`es Peraldi-Frati, and Fr\'ed\'eric Mallet}
\IEEEauthorblockA{\textit{Universit\'e C\^ote d'Azur, Inria, CNRS, I3S, Kairos}\\
Sophia Antipolis, France\\
maksym.labzhaniia@etu.univ-cotedazur.fr, julien.deantoni@inria.fr, \{marie-agnes.peraldi, frederic.mallet\}@univ-cotedazur.fr}%
}

\begin{document}
\maketitle

\begin{abstract}
Scenario-based validation is essential for automated road vehicles but cannot cover the full variability of open-world traffic.
We present MakSafe, a formal spatiotemporal framework for specifying and checking road vehicle safety rules through semantic zones, action-based forecasting, and trace-tree reasoning, avoiding uniform time discretization.
We formalize the road network, vehicle states, and occupancy zones, then build executable rules for front-vehicle identification and time-to-collision (TTC) verification.
The evaluation combines a trace-dependent case study with a scalability benchmark over traffic density ($N$), horizon ($H$), and branch factor ($B$).
MakSafe yields branch-specific verdicts, deriving safe and unsafe outcomes from the same initial scene, and re-evaluates a scenario after a control update.
A single rule check stays well below a millisecond across the reported sweep, and longer horizons slow the TTC check.
\end{abstract}


\section{Introduction}
Advanced Driver Assistance Systems (ADAS) and increasingly autonomous driving functionalities require rigorous mechanisms for specifying and assessing safety.
Standards such as ISO~21448 SOTIF and ISO~26262 define development and validation processes for automotive systems, yet the operational validation of driving behaviour remains difficult due to the combinatorial complexity of traffic situations.

Scenario-based validation is today the dominant industrial practice.
However, a scenario only captures a finite sequence of scenes (snapshots of the environment and all vehicle states) and consequently, even large scenario libraries cannot guarantee coverage of all future traffic evolutions that may emerge from a given scene.
This limitation is especially problematic for systems that must reason about future evolution and interactions among dynamic agents.

Safety rules constitute a complementary abstraction.
Instead of validating isolated scenarios, safety rules specify general constraints over acceptable vehicle behaviour.
Typical examples include maintaining a minimum time headway, ensuring safe lane-change distances, or preserving collision-free occupancy over a given horizon.

Many existing formal approaches rely on discretized temporal traces and evaluate safety at uniformly sampled timestamps.
While effective, such approaches can incur large computational overheads and may require fine-grained sampling to capture semantically relevant transitions between sampling instants.
In this paper, we present MakSafe, a spatiotemporal semantic framework for specification and compliance checking of road vehicle safety rules.
MakSafe relies on semantic zones representing relevant spatial occupancy over road networks and performs safety reasoning around decision points induced by actions rather than uniformly discretizing time.
The framework supports prospective safety assessment, enabling interpretable, machine-checkable verdicts for predicted future behaviour.

The main contributions of this paper are:

\begin{itemize}[leftmargin=*]
  \item A formal semantic framework for representing road networks, vehicles, actions, and spatial zones;
  \item A trace-tree forecasting mechanism that supports action-dependent prospective safety reasoning and branch-specific verdict generation;
  \item A Haskell implementation with two evaluations: (i) a lane-end time-to-collision (TTC) case study with trace-specific verdicts, and (ii) a scalability benchmark.
\end{itemize}

We organize the remainder of the paper as follows.
Section~\ref{sec:background} summarizes the background and related work.
Section~\ref{sec:approach} introduces the formal foundations and verification procedures of MakSafe.
Section~\ref{sec:assessment} presents the lane-end TTC case study and trace-dependent evaluation results.
Section~\ref{sec:discussion} discusses implications and limitations.
Section~\ref{sec:conclusion} concludes.

\section{Background and Related Work}
\label{sec:background}
% Scenario-based validation remains central in industrial autonomous-driving V\&V, but cannot exhaustively cover open-world traffic variability~\cite{geoscenario2019,sceml2020,menzel2018,babikian2024}. In parallel, formal traffic-rule approaches and runtime monitors improved machine-interpretable compliance checking~\cite{rizaldi2015,rizaldi2017,maierhofer2020,esterle2020,censi2019,collin2020,lawbreaker2022}. Hybrid and temporal-logic methods provide strong mathematical foundations~\cite{alur1994,henzinger1996,uppaal2006,maler2004monitoring}, and lane-centric road abstractions are widely used in practice~\cite{lanelets2014,opendrive}.

% MakSafe complements these lines of work by combining semantic occupancy zones with action-based forecasting. Compared with uniform time discretization, MakSafe evaluates rule compliance at behaviourally meaningful decision points within rule-derived horizons. The resulting analysis is trace-specific, interpretable, and directly executable.

%\subsection{Scenario-Based Validation and Testing}

Frameworks such as GeoScenario~\cite{geoscenario2019} and SceML~\cite{sceml2020}, together with industrial standards such as OpenSCENARIO~\cite{openscenario}, support the structured specification of driving situations and testing workflows.
Menzel \emph{et al.}~\cite{menzel2018} discussed the central role of scenarios for development and validation of automated vehicles.
More recently, Babikian \emph{et al.}~\cite{babikian2024} proposed optimization-based concretization mechanisms for abstract traffic-scene specifications.
While these approaches are effective for test generation and simulation, they primarily focus on concrete scenario instantiation rather than semantic reasoning about general traffic-safety properties.

%\subsection{Formalization of Traffic Rules}

Several works investigated machine-interpretable formalization of traffic laws.
Rizaldi and Althoff~\cite{rizaldi2015} formalized traffic rules for accountability of autonomous vehicles, while Rizaldi \emph{et al.}~\cite{rizaldi2017} introduced a formally verified monitoring framework in Isabelle/HOL.
Maierhofer \emph{et al.}~\cite{maierhofer2020} expressed interstate traffic rules using temporal logic, and Esterle \emph{et al.}~\cite{esterle2020} investigated machine-interpretable representations of traffic regulations.
Rulebook-oriented approaches~\cite{censi2019,collin2020} define prioritized behavioural constraints for autonomous-driving decision making.
LawBreaker~\cite{lawbreaker2022} combines traffic-law specification with fuzz testing for autonomous vehicles.
These works provide important formal foundations for traffic-law reasoning.
However, many evaluate over discretized traces and check logical compliance, without modelling continuous spatial occupancy.

%\subsection{Hybrid Systems and Temporal Logic}

Formal verification of cyber-physical systems frequently relies on temporal logic and hybrid-system models.
Timed and hybrid automata~\cite{alur1994,henzinger1996} provide rigorous models combining continuous dynamics and discrete transitions, while tools such as UPPAAL~\cite{uppaal2006} support model checking of timed systems.
Signal Temporal Logic (STL)~\cite{maler2004monitoring} expresses temporal properties over continuous signals and has become popular for autonomous-driving verification.
MakSafe shares conceptual similarities with hybrid-system approaches since vehicle evolution combines continuous motion and discrete behavioural actions.
However, unlike classical hybrid automata, MakSafe centres its reasoning on semantic spatial zones and action-oriented trace exploration rather than global state-space discretization.

%\subsection{Road-Network and Spatial Representations}

Road topology representation is another important aspect of autonomous-driving verification.
Lanelets~\cite{lanelets2014} provide an efficient representation for structured road geometries and are common in autonomous-driving systems.
MakSafe currently relies on semantic lane segments and connectivity relations.
The framework is compatible with richer geometric representations and future work will investigate integration with lanelet-based infrastructures and OpenDRIVE~\cite{opendrive} road descriptions.

%\subsection{Positioning of MakSafe}

% MakSafe differs from existing approaches in three principal aspects:

% \begin{itemize}[leftmargin=*]
%     \item it combines semantic spatial occupancy with prospective behavioural reasoning;

%     \item it performs verification around semantically meaningful action points rather than uniformly discretized timestamps;

%     \item it provides executable semantic reasoning through a formally defined Haskell implementation.
% \end{itemize}

% The framework is therefore positioned at the intersection of semantic traffic-rule formalization, spatiotemporal reasoning, and executable formal methods for autonomous-driving safety analysis.

MakSafe complements these lines of work by combining semantic occupancy zones with action-based forecasting.

\section{MakSafe: A Framework for Spatiotemporal Safety Reasoning}\label{sec:approach}

MakSafe is a formal spatiotemporal framework for specifying and checking road vehicle safety rules.
Reasoning about semantic zones and action-triggered decision points, rather than uniformly sampled time traces, yields interpretable verdicts and bounded exploration under rule-derived horizons.

\subsection{Static Road Network Encoding}

MakSafe models the road network as a set of lanes.
We represent a lane as
\begin{equation}
\ell = (\rho, s_0, s_1, \lambda, \delta),
\end{equation}
where $\rho \in \mathbb{N}$ is the road identifier, $s_0, s_1 \in \mathbb{R}$ are longitudinal start and end coordinates with $s_0 < s_1$, $\lambda \in \mathbb{Z}$ is the lateral lane index, and $\delta \in \{\mathsf{Forward},\mathsf{Backward}\}$ is the driving direction.
We denote the set of all lanes $\mathcal{L}$.

A lane segment is
\begin{equation}
\mathsf{seg} = (\ell, x_1, x_2), \quad x_1,x_2 \in [s_0,s_1],\ x_1 \le x_2.
\end{equation}

Directed links model connectivity:
\begin{equation}
 e = (\ell_1, \ell_2),
\end{equation}
where $\ell_1, \ell_2 \in \mathcal{L}$ are the source and target lanes.
We denote the set of all links $\mathcal{E}$.
A lane graph $G = (\mathcal{L}, \mathcal{E})$ is the topological layer, the substrate for zone propagation and reachability computation.

\subsection{Participants and Zones}

We represent a vehicle as
\begin{equation}
  v = (i, \mathit{name}, Len, u_{\max}),
\end{equation}
where $i$ is the vehicle identifier, $\mathit{name}$ is its name, $Len \in \mathbb{R}_{>0}$ is the longitudinal length, and $u_{\max} \in \mathbb{R}_{\ge 0}$ is the maximum speed.
We define a position as $p=(\ell,x)$ where $\ell \in \mathcal{L}$ and $x \in [s_0, s_1]$ is a longitudinal coordinate.
A dynamic vehicle state is
\begin{equation}
\sigma = (p, u, a, \delta_v, A, L),
\end{equation}
where $p=(\ell,x)$ is the current reference position (the longitudinal centre of the vehicle) on lane $\ell$, $u \in [0, u_{\max}]$ is the current speed, $a$ is the acceleration, $\delta_v$ is the movement direction, $A$ is the set of lanes being assumed (entered but reference point not yet fully inside), and $L$ is the set of lanes being left (exited but part of vehicle body remains).
In the current implementation, prospective propagation inside one action interval, a maximal action-free span of a forecast trace, uses constant longitudinal acceleration (possibly zero), with speed saturation at physical bounds.

A zone is a finite set of lane segments:
\begin{equation}
Z = \{\mathsf{seg}_1,\ldots,\mathsf{seg}_n\}.
\end{equation}
We use fusion (i.e., the minimal union of overlapping or adjacent segments in a zone), union, and intersection operators over zones; we define overlap as non-empty intersection.

Let $Z_\sigma$ collect one full-lane segment for each lane the vehicle occupies in state $\sigma$: the current lane and the lanes in $A$ and $L$.
We define the collision zone, the vehicle body's footprint, as a one-dimensional metric ball of radius $Len/2$ centred at the position $p$, intersected with $Z_\sigma$:
\begin{equation}
\mathsf{CollisionZone}(v,\sigma)=\mathsf{ball}\!\left(p,\tfrac{Len}{2}\right)\cap Z_\sigma.
\end{equation}

A \textit{reachability zone} is a semantic zone comprising the positions a vehicle's shape can occupy under the assumed forecast, starting from a given state and extending over a specified time horizon.
Over the horizon, we assume the vehicle preserves its acceleration profile and performs no optional manoeuvres, while forced linkage events fire as they become feasible.
\begin{equation}
\mathsf{ReachZone}(v,\sigma,t)=\mathsf{fuse}\!\left(\bigcup_{0\le\tau\le t}\mkern-12mu\mathsf{CollisionZone}(v,\sigma_\tau)\right)\mkern-7mu,
\end{equation}
the footprint swept as the vehicle advances over the horizon, where the union also spans each $\sigma_\tau\in\mathit{forecast}(\sigma,\tau)$, the states the forecast admits at time $\tau$ by advancing $\sigma$ along the lane graph by the travelled distance $d(\tau)$, turning multi-valued at links, and updating the lanes $Z_{\sigma_\tau}$ covers.
In the current implementation, $d(t)$ is acceleration-aware and piecewise: while acceleration is active,
\begin{equation}
d(t)=u\,t+\tfrac{1}{2}a t^2,
\end{equation}
and after saturation (full stop for braking or $u_{\max}$ for acceleration), propagation continues at the corresponding constant limiting speed.

MakSafe computes reachability zones without sampling intermediate instants, propagating the vehicle's reference position along the lane graph's feasible links by the distance travelled over the horizon and accounting for the vehicle length.

A \emph{scene} $q$ is a snapshot of the road network and all vehicle states at a given instant.
Formally, it consists of a lane graph together with the states of all vehicles.
Scenes are the atomic units for both rule evaluation and scenario construction; a scenario is a sequence of scenes over time.

\subsection{Action Modelling and Forecasting}

Starting from a single scene, multiple future evolutions may emerge through different action choices.
Accordingly, MakSafe performs prospective safety verification by generating a branching set of future traces and evaluating safety conditions independently on each forecast branch, producing trace-specific verdicts.
MakSafe models vehicle behaviour through actions, which can be either forced or unforced.
Forced actions correspond to linkage events, while unforced actions are optional manoeuvres.
We use forced actions for lane-transition events: $\mathsf{AssumeLane}(\ell_1,\ell_2)$, $\mathsf{LeaveLane}(\ell)$, $\mathsf{ExecuteTransition}(\ell_2)$.

\begin{figure}[ht]
  \centering
  \includegraphics[width=0.86\columnwidth]{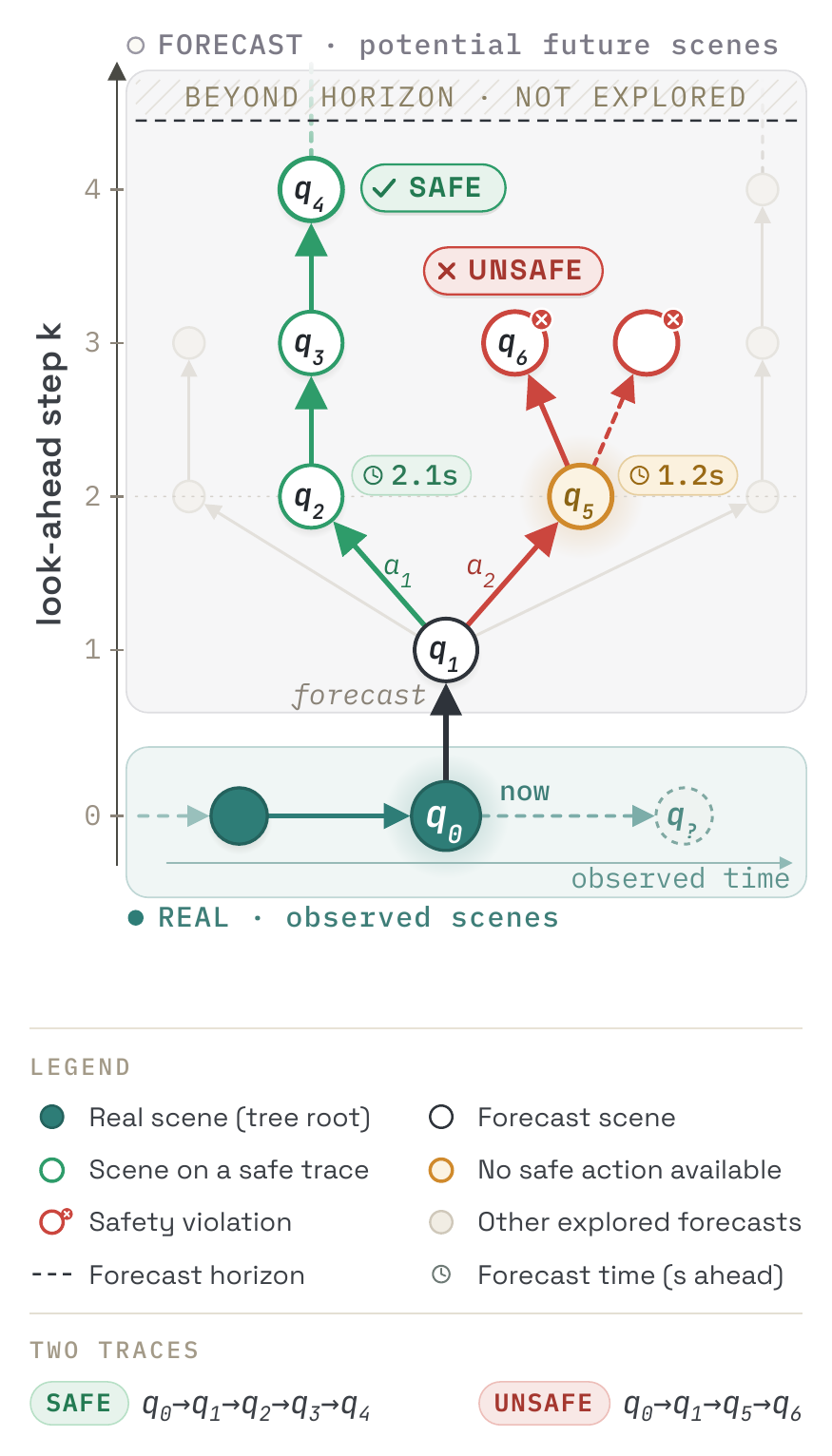}
  \caption{Trace-tree structure for prospective rule evaluation.
Observed scenes seed a forecast tree of scenes $q_i$ with actions as edges; MakSafe evaluates each branch independently up to the horizon, yielding per-trace verdicts.
Forecasting advances on actions, not uniform time, so one look-ahead step $k$ occurs at different forecast times across branches ($q_2,q_5$ sit at $k{=}2$, yet $2.1$ and $1.2$~s ahead).}
  \label{fig:trace-structure}
\end{figure}

Applying an action $\alpha$ to a state $\sigma$ yields a successor state $\sigma'$.
The forecast function repeatedly applies admissible actions and stores the resulting evolutions in a trace tree, where nodes represent scenes and edges represent forecasting in time.
MakSafe explores the tree up to the rule-derived horizon, producing reachable future states and their corresponding action prefixes.
Figure~\ref{fig:trace-structure} illustrates this structure: initial scenes come from sensing or simulated environments and form the roots; forecast branches encode alternative action sequences (for example linking to a lane or another at a specific forecasted time).

At the specification level, users compose rules from these semantic primitives (zones, actions, forecast horizons, and geometric predicates) through executable Haskell combinators.
In this paper, we focus on this reasoning core rather than introducing a separate external surface grammar.

\subsection{Illustrative MakSafe Constructs}

To demonstrate how MakSafe supports the construction of safety-relevant executable definitions, we present two illustrative rules: front-vehicle identification and time-to-collision verification.
Both rules operate in a prospective mode from an initial scene (provided either by sensing or a simulator), and the framework forecasts the near-future evolution of all participants before evaluating the safety condition on each reachable path.
The output of the first rule is therefore a trace-indexed set of front vehicles, not a static vehicle list.

\subsubsection{Front-Vehicle(s) Identification}

Front-vehicle(s) identification is itself a spatiotemporal and horizon-relative rule: the answer may change with time, with the action prefix the ego vehicle takes, and with the chosen horizon.
As specified in Algorithm~\ref{algo:frontVehicleAlgo}, for each ego trace, MakSafe identifies vehicles that are (i) ahead in the current lane or (ii) the earliest opposite-direction entrants into the ego lane before horizon expiration or ego lane change.
The rule therefore reads as a tree-structured forecast: the same initial scene can yield different front-vehicle sets on different branches, with different qualifying times and even different horizon cutoffs.
Concretely, the algorithm must forecast either to test whether another vehicle may enter the ego lane before the horizon expires, or to extend the ego trace through the next reachable forward lane set when no front vehicle has qualified yet.
Operationally, only \emph{ego} actions label trace branches: MakSafe forecasts each surrounding vehicle deterministically (with constant acceleration in the current implementation), though a forecast that crosses a link still multiplies the propagated state combinations.
For each ego branch and its associated time window, the algorithm queries each surrounding vehicle to extract local predicates (current occupancy or earliest lane-entry time into the ego lane), and records the corresponding qualifying time on that same ego branch.
This ego-centric design keeps the branching factor bounded even when a forecast turns multi-valued, as in the benchmark at $H{=}6$~s: the extra states multiply zone checks within a branch, never the branch count.

\begin{algorithm}
\caption{Trace-Dependent Front Vehicle Identification for an Ego Vehicle $e$}
\label{algo:frontVehicleAlgo}
\KwIn{Scene $q$, Ego Vehicle $e$, Time Horizon $T$}
\KwOut{Set of trace-indexed front vehicles $\mathcal{F}$ with ego trace $\pi_e$ and qualifying time $t_v$}

$\mathcal{F} \gets \emptyset$\;
$Q \gets \{(\text{empty trace}, \text{current lane of }e, t=0)\}$\;

\While{$Q \neq \emptyset$}{
  $(\pi_e,\ell_e,t_e) \gets$ dequeue from $Q$\;
  \ForEach{vehicle $v \neq e$ in $q$}{
    \If{$v$ is ahead of $e$ in lane $\ell_e$ at time $t_e$}{
      Add $(v,\pi_e,t_e)$ to $\mathcal{F}$\;
    }
    \ElseIf{$v$ may enter lane $\ell_e$ from opposite direction before $T$ or before ego lane change}{
      $t_v \gets$ earliest predicted entry time of $v$ into $\ell_e$ within the current ego time window\;
      Add $(v,\pi_e,t_v)$ to $\mathcal{F}$\;
    }
  }
  \If{this node's set is empty and linkage actions can extend $\pi_e$}{
    \ForEach{next reachable forward lane $\ell'_e$}{
      \lIf{$t'_e \le T$}{Enqueue $(\pi'_e,\ell'_e,t'_e)$ into $Q$}
    }
  }
}
\Return{$\mathcal{F}$}
\end{algorithm}

\subsubsection{Formalization of the Time-to-Collision Rule}

Building on front-vehicle identification, we formalize the rule that the ego vehicle must maintain at least a two-second time-to-collision (TTC) with front vehicles.
Prospection is again central: MakSafe computes the ego reachability zone under forecasted dynamics, so the check reflects whether a collision \emph{would} occur, on a specific ego trace, within the safety horizon, not whether one is already occurring.
The horizon is closed, so an overlap arising exactly at the two-second bound already counts as a violation, which places the check on the conservative side of the threshold.

Starting from an initial global state, reachable configurations at time $T$ are
\begin{equation}
\mathsf{Reach}(q_0,T)=\prod_{i=1}^{n}\mathsf{Reach}_{v_i}(\sigma_i,T),
\end{equation}
where each $\mathsf{Reach}_{v_i}(\sigma_i,T)$ collects the states $v_i$ can reach from $\sigma_i$ within horizon $T$.
A collision occurs when some reachable configuration has overlapping zones:
\begin{equation}
\begin{aligned}
&\exists\,(\sigma'_1,\ldots,\sigma'_n)\in\mathsf{Reach}(q_0,T),\ \exists\, i \neq j:\\
&\quad\mathsf{CollisionZone}(v_i,\sigma'_i)\cap\mathsf{CollisionZone}(v_j,\sigma'_j)\neq\emptyset.
\end{aligned}
\end{equation}

As specified in Algorithm~\ref{alg:ttc-threshold}, we forecast both the ego and the front vehicle over the reduced horizon.
For a same-direction candidate we test overlap between the ego reachability zone and the front vehicle's forecast collision zones at finitely many critical instants clamped to the horizon endpoint.
Opposite-direction candidates instead intersect the two reachability zones.

\begin{algorithm}
\caption{Time-to-Collision Threshold Verification}
\label{alg:ttc-threshold}
\KwIn{Scene $q$, Ego Vehicle $e$, TTC Threshold $T$}
\KwOut{Set of violations $\mathcal{V}$ with front vehicle, ego trace, and qualifying time}

$\mathcal{V} \gets \emptyset$\;
$\mathcal{F} \gets$ FrontVehicleIdentification($q,e,T$)\;

\ForEach{front vehicle $(v,\pi_e,t_v)$ in $\mathcal{F}$}{
  $T' \gets T - t_v$\;
  $rz_e(t) \gets$ ReachZone($e,\mathit{apply}(\sigma_e,\pi_e),t$)\;
  \eIf{ego and front vehicle move in the same direction}{
    $\{t_k\} \gets$ the defined closest-approach, speed-saturation, and eventual-overtake instants, clamped to $T'$\;
    $\mathit{hit} \gets \exists\,t_k\ \exists\,\sigma'\!\in\!\mathit{forecast}(\sigma_v,t_k):\, rz_e(t_k) \cap \mathsf{CollisionZone}(v,\sigma') \neq \emptyset$\;
  }{
    $rz_v \gets$ ReachZone($v,\sigma_v,T'$)\;
    $\mathit{hit} \gets rz_e(T') \cap rz_v \neq \emptyset$\;
  }
  \lIf{$\mathit{hit}$}{Add $(v,\pi_e,t_v)$ to $\mathcal{V}$}
}
\Return{$\mathcal{V}$}
\end{algorithm}

\subsection{Complexity}

%MakSafe separates a topological layer, a geometric layer, and a dynamical layer. This supports modular modelling and reusable verification logic across scenarios.

We characterize the computational cost of rule evaluation by separating ego-branch exploration from per-vehicle query processing.

If $A_e$ is the number of admissible ego actions per node and $D$ the bounded exploration depth, ego-trace exploration is approximately
\begin{equation}
O(A_e^D).
\end{equation}
The end-to-end cost also scales with the number of surrounding vehicles $N$ and the per-vehicle query cost $C_{\text{query}}$ used for occupancy/entry-time checks, yielding $O(A_e^D\cdot N\cdot C_{\text{query}})$ in the typical ego-anchored workflow.
Two design choices bound this cost: evaluating at action points rather than at $N_t$ uniformly sampled time steps removes the time-sampling factor, and ego-only branch labels avoid the $O(\prod_i A_{v_i}^D)$ blow-up of enumerating every agent's choices whenever surrounding forecasts stay single-valued within the horizon.
The cost is coverage: MakSafe does not explore non-deterministic surrounding behaviour, so verdicts hold only under the assumed forecast.

The artifact exposes this specification model as a host-language DSL in Haskell rather than as a standalone surface language.
A dedicated end-user concrete syntax, such as a BNF-defined surface language, remains future work.

\section{Spatiotemporal Safety Assessment}
\label{sec:assessment}

\subsection{Verification Under TTC Constraints}

We evaluate MakSafe on a scenario that demonstrates its core contribution: assessing safety under spatiotemporal constraints where the trace tree determines different front vehicles for different action choices, yielding divergent safety verdicts.
Specifically, the ego vehicle approaching a lane transition must evaluate whether each lane choice violates the TTC rule, a spatiotemporal constraint requiring the ego to maintain enough distance from the front vehicle such that a collision would occur more than 2 seconds in the future (i.e., a time-to-collision above $2$~s is safe).
This assessment is trace-dependent: the choice of lane determines which vehicle becomes the front vehicle, and thus which verdict applies.
The Haskell implementation directly encodes and executes the scenario.

\subsubsection{Road Network and Initial Scene}

\begin{figure}[ht]
  \centering
  \includegraphics[width=\columnwidth]{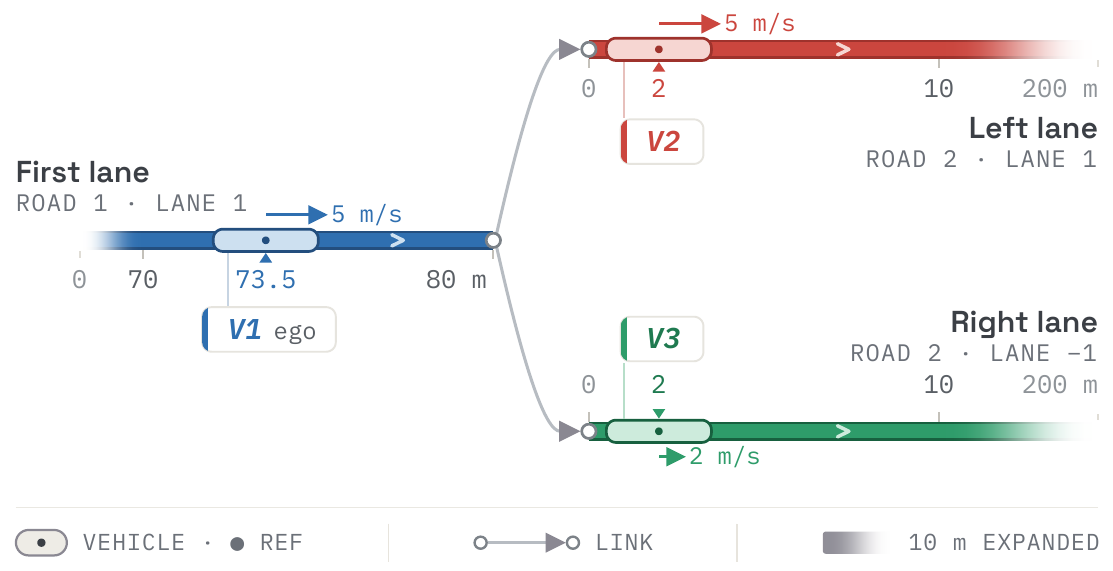}
  \caption{Road topology for the evaluation scenario.}
  \label{fig:topology}
\end{figure}

The road network, shown in Fig.~\ref{fig:topology}, consists of three lanes: the \emph{First lane}: road 1, lateral index 1, longitudinal span $[0, 80]$~m; the \emph{Left lane}: road 2, lateral index 1, longitudinal span $[0, 200]$~m; the \emph{Right lane}: road 2, lateral index $-1$, longitudinal span $[0, 200]$~m.
Two links connect the first lane to the left and right lanes respectively.

\begin{figure}[ht]
  \centering
  \includegraphics[width=0.73\columnwidth]{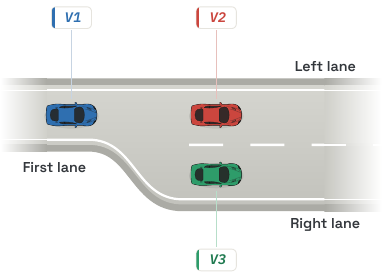}
  \caption{Initial scene: ego V1 (blue) in the first lane, V2 (red) in the left lane, V3 (green) in the right lane.}
  \label{fig:scene}
\end{figure}

The initial scene (Fig.~\ref{fig:scene}) contains three vehicles, each with a length of 3~m.
The ego vehicle V1 starts at $73.5$~m in the first lane and travels at $5$~m/s.
Because it is close to the lane end, it must imminently take one of the two available links.
Vehicle V2 sits at $2$~m in the left successor lane and also travels at $5$~m/s, while vehicle V3 lies at $2$~m in the right successor lane and travels at $2$~m/s.

\subsubsection{Identification of Front Vehicles}

\begin{figure}[ht]
  \centering
  \includegraphics[width=0.88\columnwidth]{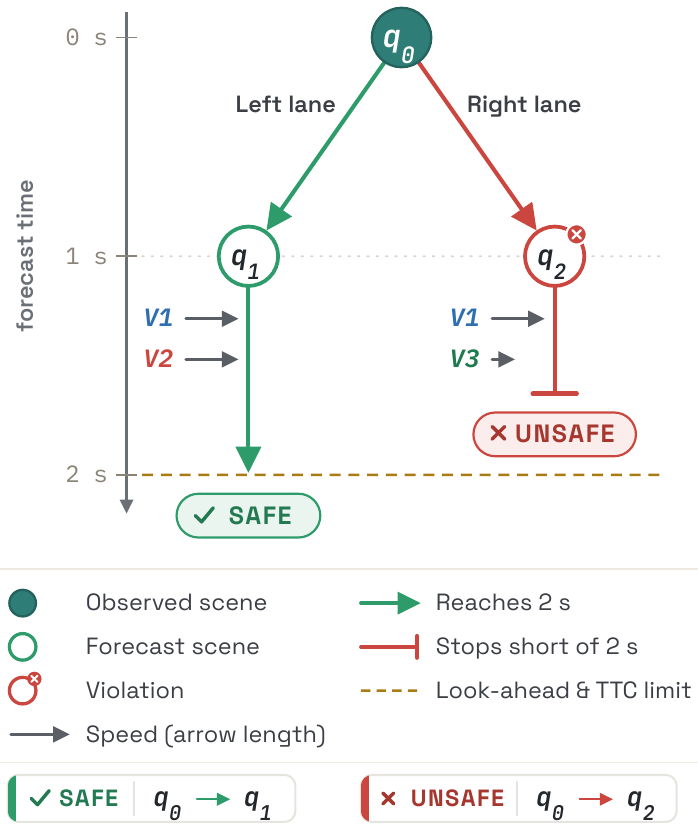}
  \caption{Trace tree for the TTC rule.
Assuming the left lane is safe ($\pi: q_0\to q_1$); assuming the right lane violates TTC with V3 ($\pi: q_0\to q_2$).}
  \label{fig:traces}
\end{figure}

A key property of MakSafe is that the safety verdict is \emph{trace-dependent}, not a global Boolean.
The trace tree, rooted at the initial scene, branches according to the ego's admissible actions.
Here, the ego has two lane-assumption actions at $t = 1.0$~s: assume the left lane or assume the right lane.
As illustrated in Fig.~\ref{fig:traces}, each action branch identifies a different front vehicle.
The framework then evaluates the TTC rule independently for each trace, resulting in opposing verdicts.

The front-vehicle identification algorithm returns both V2 and V3 as front vehicles, each associated with the trace that leads the ego toward its lane.

Since the ego speed is $5$~m/s and V3 moves at only $2$~m/s, under a two-second horizon the ego reachability zone overlaps V3's collision zone if the ego takes the right-lane transition, giving a TTC violation.
The left-lane assumption is safe because the relative longitudinal speed between the ego and V2 is zero, so no TTC violation occurs within the horizon.

% \begin{table}[ht]
% \centering
% \caption{Trace-Specific Verification Outcomes (initial scene, $T=2$~s)}
% \label{tab:results}
% \begin{tabular}{llcc}
% \toprule
% Trace & Front Vehicle & TTC $\le 2$~s & Verdict \\
% \midrule
% $\pi_1$: assume left lane  & V2 & No  & Safe   \\
% $\pi_2$: assume right lane & V3 & Yes & Unsafe \\
% \bottomrule
% \end{tabular}
% \end{table}

\subsubsection{Scenario-Level Check}

Beyond a single scene, MakSafe evaluates safety over a \emph{scenario}, i.e.\ a timed sequence of scenes.
We define a two-scene scenario: a control update decelerates the ego from $5$~m/s to $2$~m/s.
After this update, MakSafe re-evaluates the rule from the same reference position: the slower ego no longer reaches the lane transition within the two-second horizon, so no front vehicle qualifies and MakSafe reports no violation.
The deceleration restores safety.

\subsection{Performance Evaluation}

To complement the qualitative case study, we evaluate runtime with the benchmark harness in the released artifact.
The benchmark characterizes implementation-level computational behaviour rather than establishing asymptotic complexity bounds for all rule classes.
For each configuration, we report front-vehicle identification time, TTC-check time, and their sum (\emph{total} time), where $t_{\mathrm{total}} = t_{\mathrm{frontVehicle}} + t_{\mathrm{TTC}}$.
The TTC check identifies front vehicles internally, so $t_{\mathrm{TTC}}$ already covers that work and $t_{\mathrm{total}}$ upper-bounds a single end-to-end check rather than measuring one.
Table~\ref{tab:perf} reports the recorded baseline's means; the README relates them to a rerun.

The measured campaign covers $N\in\{10,25,50\}$ surrounding vehicles, horizons $H\in\{2,6\}$~s, and branch factors $B\in\{2,4\}$.
The benchmark places surrounding vehicles deterministically from the run seed (lane, offset, speed in $2$--$7$~m/s, acceleration in $[-0.5,0.5]$~m/s$^2$) and fixes the ego at $610$~m on a $650$~m lane at $4$~m/s with acceleration $0.5$~m/s$^2$.
It reports one \texttt{System.CPUTime} delta spanning $80$ inner iterations, then averages over $5$ repeats, forcing each result, including its rendered textual report, to weak-head normal form through an integer checksum.

The benchmark generator parameterizes topology by $B$: one approach lane splits into $B$ split lanes, then into $B$ branch lanes, which merge back into a downstream corridor of three consecutive lanes.
The resulting synthetic network contains $2B+4$ lanes and $3B+2$ links.
The ego reaches no link within either horizon, yet $B$ raises the TTC-check timings in Table~\ref{tab:perf}: it widens the admissible-link scan, and a front vehicle reaching a link within $H$ forecasts one state per outgoing link.

\begin{table}[ht]
\centering
\caption{Performance Results (Aggregated Means)}
\label{tab:perf}
\begin{tabular}{cccccc}
\toprule
$N$ & $B$ & $H$ (s) & frontVehicle (ms) & TTC check (ms) & total (ms) \\
\midrule
10 & 2 & 2 & 0.787 & 1.049 & 1.836 \\
10 & 2 & 6 & 0.583 & 1.455 & 2.038 \\
10 & 4 & 2 & 0.755 & 1.118 & 1.873 \\
10 & 4 & 6 & 0.760 & 1.877 & 2.638 \\
\midrule
25 & 2 & 2 & 0.803 & 1.297 & 2.101 \\
25 & 2 & 6 & 0.808 & 1.801 & 2.610 \\
25 & 4 & 2 & 0.979 & 1.663 & 2.641 \\
25 & 4 & 6 & 0.887 & 3.087 & 3.974 \\
\midrule
50 & 2 & 2 & 1.242 & 1.840 & 3.082 \\
50 & 2 & 6 & 1.249 & 2.260 & 3.509 \\
50 & 4 & 2 & 1.322 & 2.257 & 3.579 \\
50 & 4 & 6 & 1.306 & 3.562 & 4.869 \\
\bottomrule
\end{tabular}
\end{table}

Overall, runtimes remain within a few milliseconds across all reported configurations (about $1.8$ to $4.9$~ms in Table~\ref{tab:perf}), which bounds one end-to-end check at roughly $23$ to $61\,\mu$s.
Total time generally increases with $N$ and with horizon $H$, with the largest mean in this sweep at $(N,B,H)=(50,4,6)$.

A key point is that runtime is not strictly monotone in each individual component.
Front-vehicle identification performs identical work at both horizons in this sweep, so its variation across $H$ stays within run-to-run noise, while the TTC check grows with the horizon and dominates the total trend.

The evaluation in this paper uses the open-source implementation of MakSafe, available at~\cite{labzhaniia2026maksafe}, with all code, scenarios, and instructions for reproducing the results.

\section{Discussion}\label{sec:discussion}

MakSafe separates the topological (lanes, links, connectivity), geometric (positions, zones, reachability), and dynamical (actions, evolution over time) layers, enabling modularity and reuse of abstract definitions across traffic systems.

MakSafe evaluates safety at semantic action points rather than at uniformly sampled timestamps, reasoning over the spatial reachability zones of Section~\ref{sec:approach}.
We design the check to be conservative for the modelled longitudinal dynamics under the stated assumptions: constant-acceleration forecasting of surrounding traffic and conflicts encoded through shared lane segments.
Verdicts are thus relative to this forecast, not a guarantee over all possible evolutions, and a machine-checked soundness statement remains future work, including a proof that the computed zones match their definitions in Section~\ref{sec:approach}.

The framework does not yet model continuous acceleration variation within an action interval, jerk-bounded profiles, curved geometry, or probabilistic prediction.

\section{Conclusion}\label{sec:conclusion}

This paper presented MakSafe, a spatiotemporal semantic framework for specifying and checking road vehicle safety rules.
The framework combines semantic zones, trace-tree forecasting, and action-based reasoning to support machine-checkable prospective safety analysis without uniform time discretization.
Our evaluation supports two conclusions.
First, the case study confirms interpretability: MakSafe identifies safe and unsafe branches from the same scene and localizes the triggering action sequence.
Second, the performance study (Table~\ref{tab:perf}) places a single check in the tens of microseconds across the reported configurations, with the TTC check growing as the horizon does.

\section*{Acknowledgment}
This work has been supported by the French government, through the France 2030 investment plan managed by the Agence Nationale de la Recherche, as part of the ``UCA DS4H'' project, reference ANR-17-EURE-0004.

% grouped \centering: a center environment's \topsep would push the logo onto the reference page
{\centering
\includegraphics[width=\linewidth]{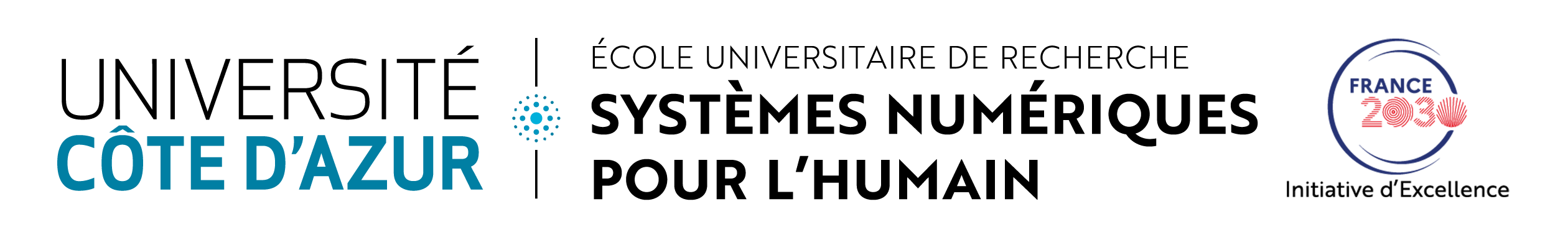}
\par}

% Column break inside the reference list, so the final page balances.
\IEEEtriggeratref{10}
\bibliographystyle{IEEEtran}
\bibliography{bibtex}

\end{document}